\documentclass{article}

\usepackage[preprint]{neurips_2020}

\usepackage[utf8]{inputenc} % allow utf-8 input
\usepackage[T1]{fontenc}    % use 8-bit T1 fonts
\usepackage{hyperref}       % hyperlinks
\usepackage{url}            % simple URL typesetting
\usepackage{booktabs}       % professional-quality tables
\usepackage{amsfonts}       % blackboard math symbols
\usepackage{nicefrac}       % compact symbols for 1/2, etc.
\usepackage{microtype}      % microtypography
\usepackage{booktabs}
\usepackage[noend]{algpseudocode}
\usepackage{algorithmicx,algorithm}
\usepackage{graphicx}
\usepackage{subfigure}
\usepackage{geometry}
\usepackage{color}
\title{RTFN: Robust Temporal Feature Network}

\author{
  Zhiwen Xiao  \\
        School of Information Science and Technology, \\
        Southwest Jiaotong University, Chengdu\\       
         \texttt{xiao1994zw@163.com} \\
   \And
     Xin Xu \\
        School of Computer Technology Science, \\
        China University of Mining and Technology\\
        \texttt{jhsu99@163.com} \\
   \And
        Huanlai Xing \thanks{Corresponding Author} \\
        School of Information Science and Technology, \\
        Southwest Jiaotong University, Chengdu\\
        \texttt{hxx@home.swjtu.edu.cn} \\
        \And
         Juan Chen\\
        School of Information Science and Technology, \\
        Southwest Jiaotong University, Chengdu\\
        \texttt{jchen@my.swjtu.edu.cn} \\
}

\begin{document}

\maketitle

\begin{abstract}
Time series analysis plays a vital role in various applications, for instance, healthcare, weather prediction, disaster forecast, etc. However, to obtain sufficient shapelets by a feature network is still challenging. To this end, we propose a novel robust temporal feature network (RTFN) that contains temporal feature networks and attentional LSTM networks. The temporal feature networks are built to extract basic features from input data while the attentional LSTM networks are devised to capture complicated shapelets and relationships to enrich features. In experiments, we embed RTFN into supervised structure as a feature extraction network and into unsupervised clustering as an encoder, respectively. The results show that the RTFN-based supervised structure is a winner of 40 out of 85 datasets and the RTFN-based unsupervised clustering performs the best on 4 out of 11 datasets in the UCR2018 archive.

\end{abstract}

\section{Introduction}

In the real world, time series data are collected from various domains, such as weather forecast, human heart record and animal trajectory [1]. How to make full use of the data in real-world applications is crucial, which depends on how well features are extracted. Recently, sufficient feature extraction has become a critical challenge, which is also a basis for time series classification [2].

In the past two decades, there are primarily two solutions to address the challenge above, including  traditional approaches and deep learning algorithms [3]. Compared with the former, deep learning has been reported to achieve better performance in feature extraction. In particular, attention networks usually achieve excellent performance in time series classification, because attention mechanism can relate different positions of a sequence in order to derive sensitive representations and relationships [4]. Besides, to embed long short term memory network (LSTM) into temporal networks is able to explore long- and short-period connections and relationships among data, enhancing feature extraction and thus facilitating the processing of subsequent tasks [5][6][7]. In short, attention and LSTM have been regarded as promising structures in the area of time series processing.

To facilitate sufficient feature extraction, this paper proposes a novel robust temporal feature network (RTFN), which consists of temporal feature networks and attentional LSTM networks. The temporal feature networks are based on residual networks [8] and multi-head convolution neural networks, which are responsible for extracting basic temporal features. In each attentional LSTM network, LSTM is embedded into an attentional structure for the purpose of efficiently mining the hidden shapelets and relationships that are usually ignored by the temporal networks. In the experiments, we embed RTFN into a supervised structure, and test it on 85 standard datasets in the UCR2018 archive, where the RTFN-based classification algorithm outperforms a number of state-of-the-art supervised algorithms on 40 datasets. Moreover, we also use RTFN in a simple unsupervised representation clustering as an encoder. The RTFN-based unsupervised clustering is a winner of 4 out of 11 datasets in the UCR2018 archive, compared with other unsupervised temporal clustering algorithms.

The rest of the paper is organized as follows. Section 2 reviews the attention and LSTM techniques. Section 3 introduces the overview of RTFN, the Attentional LSTM, the RTFN-based supervised structure and the RTFN-based unsupervised clustering. Experimental results and analysis are given in Section 4. In Section 5, summary and conclusions are provided.
\section{Related Work}
\label{gen_inst}

Feature extraction is essential for many tasks, such as computer vision, nature language processing, reasoning, etc. In general, robust feature networks are composed of targeted feature meta-networks, which take their own advantages to capture the existing features from domain data and explore the relationships among them. Attention and LSTM are two types of well recognized meta-networks in recent years, promoting the rapid development in a variety of application domains.

\subsection{Attention}
The attention mechanism aims at mapping a given query and a set of key-value pairs to an output, where the output is a weighted sum of the values. For each value, its associated weight is calculated based on a compatibility function of the query with the corresponding key [10]. For any input data, attention can mine different positions to derive the representation of input data.

Just like our eyes, the attention mechanism focuses on the sensitive information we look for, capture the areas we are interested in, and discover rich features and relationships among them. Its use is to discover the hidden information and relationships that classical networks are unaware of, which helps to further complement the time-series and semantic features. Attention has been applied successfully in various areas, such as object detection [11], image caption [12], activity recognition [13], speech recognition [14], question answering [15] and software-defined networking [16].

\subsection{LSTM}

LSTM is a widely used recurrent neural network. For a sequential model, LSTM can explore the features of input sequence by using specific internal units, and make use of these features to mine more underlying information and relationships from the internal structure [17]. Besides, the internal memory units in LSTM store the new features transmitted to the next processing units. In this way, when the features being transmitted are forgotten, they can be recovered by those stored in the internal memory units. Hence, LSTM can obtain abundant features from input data. Recently, LSTM has been incorporated into various network structures for addressing time-series tasks, for instance, ERA-LSTM [5], FCN-LSTM [6] and multivariate FCN-LSTM [18].

Motivated by the research related to the attention mechanism and LSTM, we design a robust temporal feature network that integrates LSTM into an attentional learning framework, which is named as attentional LSTM network. By taking advantages of the two techniques, the proposed network is able to not only explore long and short term temporal features, but also focus on different positions of a sequence to capture the shapelets hiding in the input data.

\section{RTFN}
\label{headings}

This section first overviews the structure of RTFN, and then describes the attentional LSTM network in detail. In the end, the RTFN-based supervised structure and unsupervised clustering are introduced.
% \usepackage{booktabs}

\subsection{Overview}

The structure of RTFN is shown in Fig. 1. It primarily consists of temporal feature networks and attentional LSTM networks. The former is based on residual networks and multi-head convolution neural networks, responsible for capturing basic temporal features from the input time-series data. Especially, in the temporal feature networks, we place self-attention networks [10] between the two multi-head convolutional neural networks to relate various positions of the shapelets that the first multi-head convolutional neural networks obtain, which enriches temporal features. The latter pays attention to digging out the complex temporal relationships from the input, which helps to compensate for the loss of features that the temporal feature networks may not capture. Therefore, combining the two types of networks together promotes the abundance of both the basic and in-depth temporal features. In this paper, RTFN is embedded into supervised structure first and unsupervised clustering later, in the context of time series classification.

\begin{figure}[htbp]
  \centering
\includegraphics[height=5.6cm,width=7.6cm]{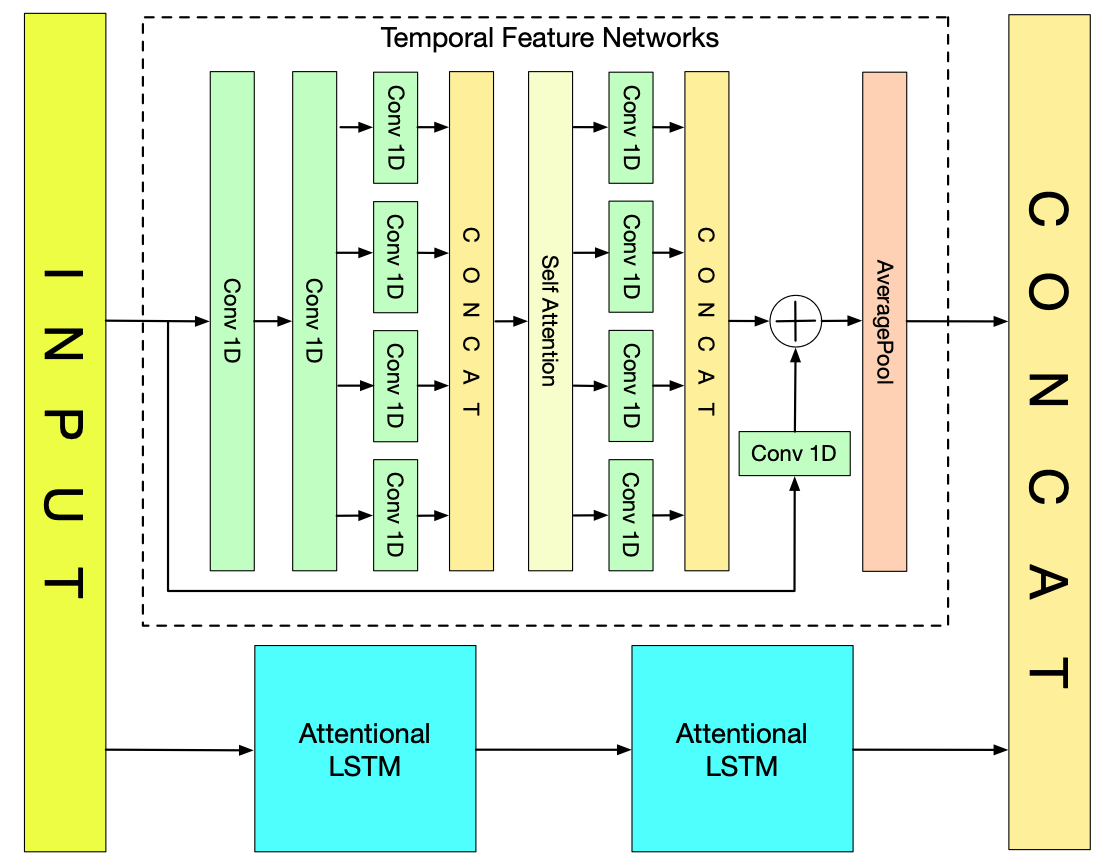}%fig2文件夹下的xbee.esp图片，
\caption{The structure of RTFN.}
\end{figure}

\subsection{Attentional LSTM}

Motivated by the original attention mechanism, we propose an innovative Attentional LSTM to concentrate on mutual temporal relationships. Its details are shown in Fig. 2.

\begin{figure}[h]
  \centering
\includegraphics[height=4cm,width=6cm]{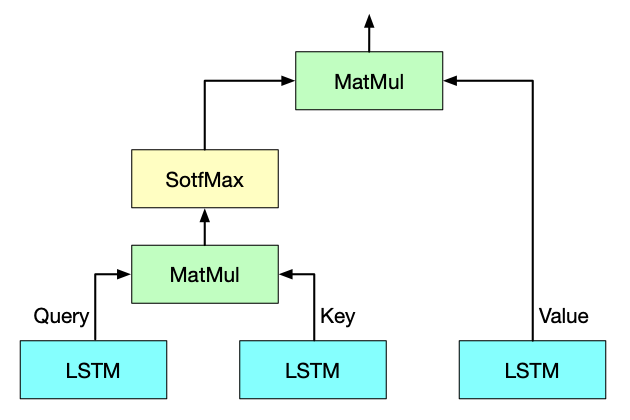}%fig2文件夹下的xbee.esp图片，
\caption{Attentional LSTM.}
\end{figure}
An attentional LSTM network is a hybridized network that incorporates LSTM networks into an attentional structure. In this network, a temporal query and a set of key-value pairs are mapped to an output, where the query, key, and value are vectors obtained from the feature extraction by the LSTM networks. The output is defined as a weighted sum of the values with sufficient shapelets and relationships. In this paper, each value is obtained by a compatibility function that helps to mine the hidden relationships between a query and its corresponding key that already carry basic features, thus strengthening the robustness of the attentional LSTM network. The calculation of the attentional LSTM network is defined in Eq. (1).

$$ ALSTM(f_{Q},f_{K},f_{V}) = softmax(f_{Q} \cdot f_{K}^{T})\cdot f_{V} \eqno{(1)}$$

where $f_{Q},f_{K}$ and $f_{V}$ represent the matrixes of the queries, keys and values, respectively.

After the sequential data go through the LSTM networks, the queries, keys and values carry sufficient long- and short-term features as input, which are then used to capture the in-depth shapelets and relationships hiding behind as output.

\subsection{The RTFN-based Supervised Structure}
The RTFN-based supervised structure is shown in Fig. 3, where a dropout layer and a fully-connected layer are cascaded to the output of RTFN. To be specific, we introduce the dropout layer to avoid overfitting during the training process. The fully-connected layer performs as the classifier of the RTFN-based supervised structure. The reason we simply use the dropout and fully-connected layers is that the features extracted by RTFN are sufficiently good and thus a complicated classifier network is not necessary. Like other commonly used supervised algorithms, we use the cross entropy function to compute the difference between the prediction result and ground truth label, shown as Eq. (2).

$$ loss_{super} =  - \frac {1}{n} \sum_{i=1}^{n} (\widehat{Y}_{i}^{train} log(p_{i})) \eqno{(2)}$$

where $\widehat{Y}_{i}^{train}$ represents the ground truth label of the $i$-th time-series class, and $p_{i}$ $\{ i=1,2,...,n\}$ is the prediction output of the RTFN-based supervised structure.

\subsection{The RTFN-based Unsupervised Clustering}

The RTFN-based unsupervised clustering is based on a widely adopted auto-encoder unsupervised architecture, as shown in Fig. 4. To be specific, it is primarily composed of RTFN as the encoder, a decoder and a K-means algorithm [19]. RTFN is responsible for obtaining basic and in-depth shapelets from the input data as many as possible. The decoder is made up of four fully-connected layers, helping to reconstruct the features captured by RTFN. Besides, the K-means algorithm acts as unsupervised classifier.

Different from that the total loss takes k-means loss into account in [20][21][22], the loss of the RTFN-based unsupervised clustering simply depends on a reconstruction loss defined in Eq. (3).

$$ loss_{rec} =   \frac {1}{n} \sum_{i=1}^{n} (\breve{X}_{i}^{train} - X_{i}^{rec})^{2} \eqno{(3)}$$

where $\breve{X}_{i}^{train}$ is the input data and $ X_{i}^{rec}$ represents the decoder output.

\begin{figure}[htbp]
\centering
\begin{minipage}[t]{0.48\textwidth}
\centering
\includegraphics[height=4cm,width=4.72cm]{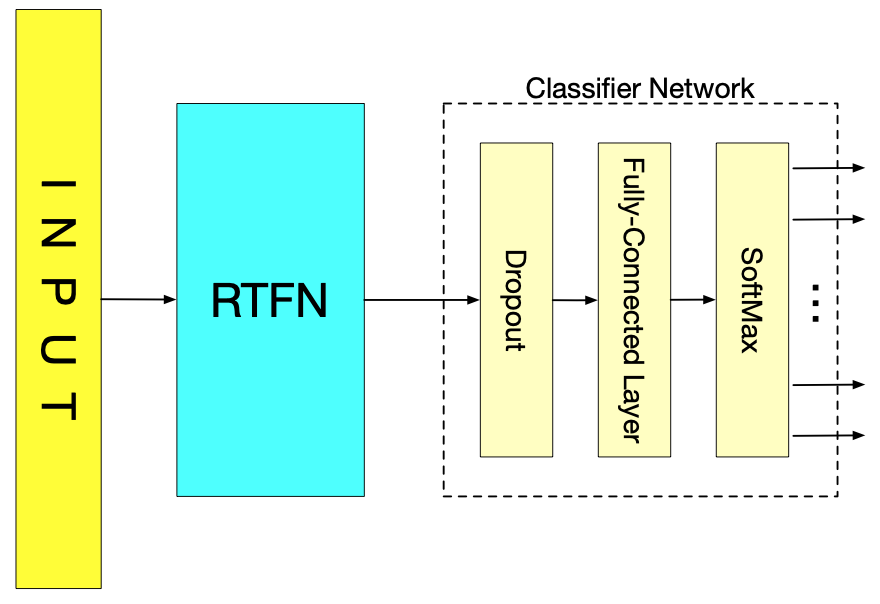}
\label{fig_22}
\caption{RTFN-based supervised structure.}
\end{minipage}
\begin{minipage}[t]{0.48\textwidth}
\centering
\includegraphics[height=4cm,width=6.2cm]{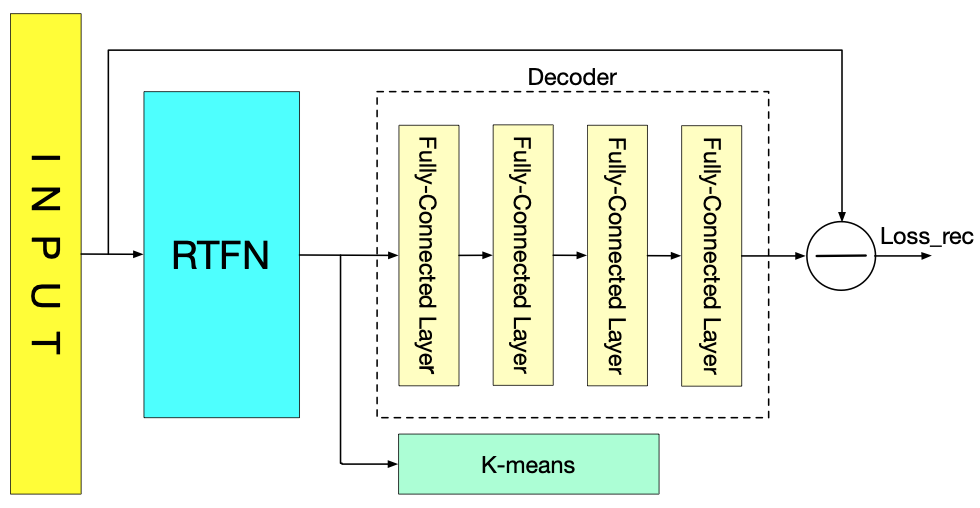}
\caption{RTFN-based unsupervised clustering.}
\label{fig_23}
\end{minipage}
\end{figure}

\section{Experiments}
\label{others}

This section first introduces the experimental setup, and then evaluates the RTFN-based supervised structure and unsupervised clustering, respectively.

\subsection{Experimental Setup}

The UCR 2018 archive is one of the authoritative temporal data archives, which contains 128 datasets with different lengths in a variety of application areas. We select 85 and 11 standard datasets to evaluate performance of the RTFN-based supervised structure and unsupervised clustering, respectively. All experiments are run on a computer with Ubuntu 18.04 OS, Nvidia dual-GTX 1070Ti GPU with 8GB $\times$ 2 GRAM and AMD R5 1400 CPU with 16G RAM. For encouraging scientific comparison, our code is available at \url{http://X.X.X.X}.

\subsection{Evaluation of the RTFN-based Supervised Structure}

To evaluate performance of the RTFN-based supervised structure, we compare it with a number of state-of-the-art supervised algorithms against the top-1 accuracy. We use ‘win’, ‘tie’ and ‘lose’ to rank the algorithms for comparison, i.e. on how many datasets that an algorithm performs better than, equivalent to, or worse than the others. For each algorithm, the number of the ‘best’ cases is the summation of the numbers of the corresponding ‘win’ and ‘tie’ cases.

Table 1 shows the statistical results obtained by different algorithms on the 85 selected datasets in the UCR 2018 archive. For each dataset, the existing SOTA represents the best algorithm on that dataset [4], including BOSS [23], COTE [24], DTW [25] and so on; similarly, for each dataset, the Best:lstm-fcn is the best-performance approach on that dataset, e.g. it involves LSTM-FCN, ALSTM-FCN, etc, in [6]. Note that the existing SOTA did not consider the last 20 of the 85 datasets. The original results are shown in the appendix.

\begin{table}[!htb]
  \caption{Statistical results obtained by different supervised methods.}
  \label{sample-table}
  \centering
  \resizebox{140mm}{8mm}{
  \begin{tabular}{cccccccc }
    \toprule

    Ranks   & Existing SOTA {[}4{]} & Best:lstm-fcn & Vanilla:ResNet-Transformer & ResNet-Transformer 1 {[}4{]} & ResNet-Transformer 2 {[}4{]} & ResNet-Transformer 3 {[}4{]} & Ours \\ \midrule
Best      & 22                    & 32            & 32                         & 23                           & 22                           & 27                           & \textbf{40}           \\

Win	&8	&\textbf{16} &5	&1	&2	&2	&11\\
Tie	&14	&16	&27	&22	&20	&25	&\textbf{29}\\
Lose	&43	&53	&53	&62	&63	&58	&45\\

    \bottomrule
  \end{tabular}
  }
\end{table}

It is seen that the proposed RTFN-based supervised structure performs the best in ‘tie’ and the second best in ‘win’, guaranteeing its first position in ‘best’. To be specific, ours wins in 11 cases and performs no worse than any other algorithm in 29 cases, which leads to 40 ‘best’ cases in the competition. Besides, the Best: lstm-fcn and the Vanilla: ResNet-Transormer achieve the second and third ‘best’ performance, with respect to the ‘best’ score. Especially, the former is a winner on 16 datasets, indicating its outstanding performance. On the other hand, the ResNet-Transformer 2 [4] is the worst algorithm, with 2 ‘win’ scores and 20 ‘tie’ scores only.

In addition, to further reflect the difference between the proposed structure and the others in the long time-series datasets, we select 12 out of the 85 datasets (see Table 2 for details) and show the top-1 accuracy results in Table 3. Besides, we add the RTFN-based supervised structure without the attentional LSTM (ours w/o a-LSTM) to the comparison, which helps to verify the contribution of the attentional LSTM.

\begin{table}[!htb]
  \caption{Details of the 12 long time-series datasets.}
  \label{sample-table}
  \centering
  \resizebox{49mm}{14.5mm}{
  \begin{tabular}{ccccc }
    \toprule

    Dataset&	Train&	Test&	Classes&	Length \\ \midrule
Beetlefly	&20	&20&	2	&512\\
BirdChicken&	20&	20	&2&	512\\
Car	&60	&60	&4&	577\\
Earthquakes& 322&139&2&512\\
Herring	&64	&64	&2&	512\\
Haptics	&155	&308&	5	&1092\\
Lightning2 &60&61&2&637\\
OliveOil	&30&	30&	4&	570\\
SemgHandGendeCh2	&300&	600&	2&	1500\\
SemgHandMovementCh2	&450&	450&	6&	1500\\
SemgHandSubjectCh2	&450	&450&	5&	1500\\
Rock & 20 & 50&4 & 2844\\

\bottomrule

  \end{tabular}
  }
\end{table}

\begin{table}[!htb]
  \caption{Results of the top-1 accuracy on the 12 datasets.}
  \label{sample-table}
  \centering
  \resizebox{140mm}{17mm}{
  \begin{tabular}{ccccccccc }
    \toprule

    Dataset  & Existing SOTA {[}4{]} & Best:lstm-fcn & Vanilla:ResNet-Transformer & ResNet-Transformer 1 {[}4{]} & ResNet-Transformer 2 {[}4{]} & ResNet-Transformer 3 {[}4{]} &Ours w/o a-LSTM & Ours \\ \midrule

Beetlefly	&0.95&	\textbf{1} &	\textbf{1}	&0.95	&0.95	&\textbf{1}	&0.85&	\textbf{1}\\
BirdChicken	&0.95	&0.95	&\textbf{1}	&0.9	&\textbf{1}	&0.7	&0.95&	\textbf{1}\\
Car	&0.933&	\textbf{0.966667}&	0.95&	0.883333&	0.866667&	0.3&	0.716667&	0.883333\\
Earthquakes	&0.801	&\textbf{0.81295}&	0.755396&	0.755396&	0.76259	&0.755396 &0.748201	&0.776978\\
Haptics	&0.551&	0.558442	&0.564935&	0.545455&	\textbf{0.600649}&	0.194805&	0.538961&	\textbf{0.600649}\\
Herring	&0.703&	\textbf{0.75}	&0.703125&	0.734375&	0.65625&	0.703125&	0.71875&	\textbf{0.75}\\
Lightning2&\textbf{0.8853}&	0.819672&	0.852459&	0.852459&	0.754098&	0.868852&0.819672	&0.836066\\
OliveOil	&0.9333&	0.766667&	\textbf{0.966667}&	0.9&	0.933333&	0.9&	0.866667&	\textbf{0.966667}\\
SemgHandGendeCh2	&---&	0.91&	0.866667&	0.916667&	0.848333&	0.651667&	0.796667&	\textbf{0.923333}\\
SemgHandMovementCh2	&---&	0.56	&0.513333&	0.504444&	0.391111&	0.468889&	0.595555&	\textbf{0.757778}\\
SemgHandSubjectCh2	&---	&0.873333&	0.746667&	0.74&	0.666667&	0.788889&	0.793333&	\textbf{0.897778}\\
Rock &---   &  \textbf{0.92}&0.78&\textbf{0.92} &0.82&0.76&0.62&0.88\\
\midrule
Best	&1&	5&	3&	1&	2&	1&	0&	\textbf{8}\\
Mean Accuracy	&---&	0.823978 	&0.808271& 	0.800177 &	0.770808& 	0.674302 &	0.750756 &	\textbf{0.856049} \\

    \bottomrule
  \end{tabular}
  }
\end{table}

On the one hand, if focusing on ours and ours w/o a-LSTM, one can easily see the former outperforms the latter on each dataset, as the former always obtains higher accuracy. This clearly unveils the attentional LSTM plays a non-trivial role in the performance improvement, helping our structure to gain 8 ‘best’ scores and the highest mean accuracy. This is because the embedded attentional LSTM is capable of extracting the underlying temporal shapelets and relationships that are simply ignored by the temporal feature networks.

On the other hand, the Best: lstm-fcn also achieves decent performance with respect to the ‘best’ score and mean accuracy, because its LSTM helps to extract additional features from input data to enrich the features obtained by the fcn networks. The Vanilla: ResNet-Transformer is no doubt the one with the best performance among all the transformer-based networks for comparison. The reason behind this is the embedded attention mechanism can further relate different positions of time series data and thus enhance the accuracy.

\subsection{Evaluation of the RTFN-based Unsupervised Clustering}
To evaluate performance of the RTFN-based unsupervised clustering, we compare it with a number of state-of-the-art unsupervised algorithms against three metrics, including the ‘best’ score as used in Section 4.2, the average ranking, and the rand index [26]. The average (AVG) rank scores are computed according to the average Geo-ranking approach, which measures the average difference between the accuracies of a model and the best accuracies among all models [22]. As a widely adopted performance indicator, the rand index (RI), $RI$, is defined in Eq. (4).

$$ RI =   \frac {PTP + NTP}{s(s-1)/2}  \eqno{(4)}$$

where  $PTP$ and $NTP$ are the numbers of the positive and negative time-series pairs in the clustering, respectively, and  $s$ is the dataset size.

\begin{table}[!htb]
  \caption{The RI results on 11 time series datasets.}
  \label{Comparion}
  \centering
  \resizebox{140mm}{18mm}{
  \begin{tabular}{cccccccccccccccc }
    \toprule

    Dataset&K-means{[}19{]}&UDFS{[}27{]}&NDFS{[}28{]}&RUFS{[}29{]}&RSFS{[}30{]}&KSC{[}31{]}&KDBA{[}32{]}&K-shape{[}33{]}&Ushapelet{[}34{]}&DTC{[}20{]}&USSL{[}35{]}&DEC{[}36{]}&IDEC{[}21{]}&DTCR{[}22{]}&Ours
\\

    \midrule
    Beef	&0.6719&0.6759&0.7034&0.7149&0.6975	&0.7057&0.6713&0.5402&0.6966&0.6345	&0.6966&0.5945&0.6276&\textbf{0.8046}&0.7655\\

    Car&0.6354	&0.6757	&0.6260	&0.6667	&0.6708	&0.6898	&0.6254	&0.7028	&0.6418	&0.6695	&0.7345	&0.6859	&0.6870	&\textbf{0.7501}	&0.7169\\
chlorineConCentrati	&0.5261	&0.5282	&0.5225	&0.5330	&0.5316	&0.5256	&0.53	&0.4111	&0.5318	&0.5353	&0.4997	&0.5334	&0.5350	&0.5357	&\textbf{0.5367}\\
Dist.phal.outl.correct	&0.5252	&0.5362	&0.5362	&0.5252	&0.5327	&0.5235	&0.5203	&0.5252	&0.5098	&0.5010	&0.5962	&0.5029	&0.5330	&0.6075	&\textbf{0.6095}\\
ECG200&	0.6315	&0.6533	&0.6315	&0.7018	&0.6916	&0.6315	&0.6018	&0.7018	&0.5758	&0.6018	&\textbf{0.7285}	&0.6422	&0.6233	&0.6648	&\textbf{0.7285}\\
GunPoint&	0.4971&	0.5029&	0.5102&	0.6498&	0.4994&	0.4971&	0.5420&	0.6278&	0.6278&	0.5400&	\textbf{0.7257}&	0.4981&	0.4974&	0.6398&	0.6471\\
Lighting2&	0.4966&	0.5119&	0.5373&	0.5729&	0.5269&	0.6263&	0.5119&	0.6548&	0.5192&	0.5770&	\textbf{0.6955}&	0.5311&	0.5519&	0.5913&	0.6230\\
OSULeaf	&0.5615&	0.5372&	0.5622&	0.5497&	0.5665&	0.5714&	0.5541&	0.5538&	0.5525&	0.7329&	0.6551&	0.7484&	0.7607&	\textbf{0.7739}&	0.7564\\
Prox.Phal.out.ageGr&	0.5288&	0.4997&	0.5463&	0.5780&	0.5384&	0.5305&	0.5192&	0.5617&	0.5206&	0.7430&	0.7939&	0.4263&	0.8091&	0.8091&	\textbf{0.8180}\\
Wafer&	0.4925&	0.4925&	0.5263&	0.5263&	0.4925&	0.4925&	0.4925&	0.4925&	0.4925&	0.5324&	\textbf{0.8246}&	0.5679&	0.5597&	0.7338&	0.8093\\
WordsSynonyms&	0.8775&	0.8697&	0.8760&	0.8861&	0.8817&	0.8727&	0.8159&	0.7844&	0.8230&	0.8855&	0.8540&	0.8893&	0.8947&	\textbf{0.8984}&	0.8973\\
 \midrule
  Best	&0	&0	&0	&0	&0	&0	&0	&0	&0	&0	&\textbf{4}	&0	&0	&\textbf{4}	&
 \textbf{4}\\
AVG rank	&11.22	&10.36&	8.90&	6.72&	8.45&	8.95&	11.90&	9.09&	11.09&	8.14&	4.81&	8.81&	6.77	&2.59&	\textbf{2.13}\\
AVG RI	&0.5858	&0.5894	&0.5980	&0.6277	&0.6027	&0.6061	&0.5804	&0.5960	&0.5901&	0.6321&	0.7095&	0.6018&	0.6436	&0.7099&	\textbf{0.7189}\\

    \bottomrule
  \end{tabular}
  }
\end{table}

We select 11 representative datasets from the UCR 2018 archive for performance evaluation and show the results in Table 4. Clearly, our RTFN-based unsupervised clustering achieves the best performance on 4 datasets. Meanwhile, it obtains the best AVG rank and AVG RI values, i.e. 2.13 and 0.7189. On the other hand, USSL [36] and DTCR [22] both achieve 4 ‘best’ scores, which are, to some extent, statistically equivalent to ours. This is because the pseudo-labels help to guide the training process in USSL while the mistake correction helps to facilitate the temporal reconstruction in DTCR. However, for USSL, there is no efficient mechanism to alleviate the negative impact on the training process when mistakes exist in pseudo-labels. Besides, the temporal reconstruction in DTCR is not good at mining cluster-specific representations from input data. This is why DTCR needs an additional K-means objective to its seq2seq structure.

\begin{figure}[h]
  \centering
\includegraphics[height=6cm,width=8cm]{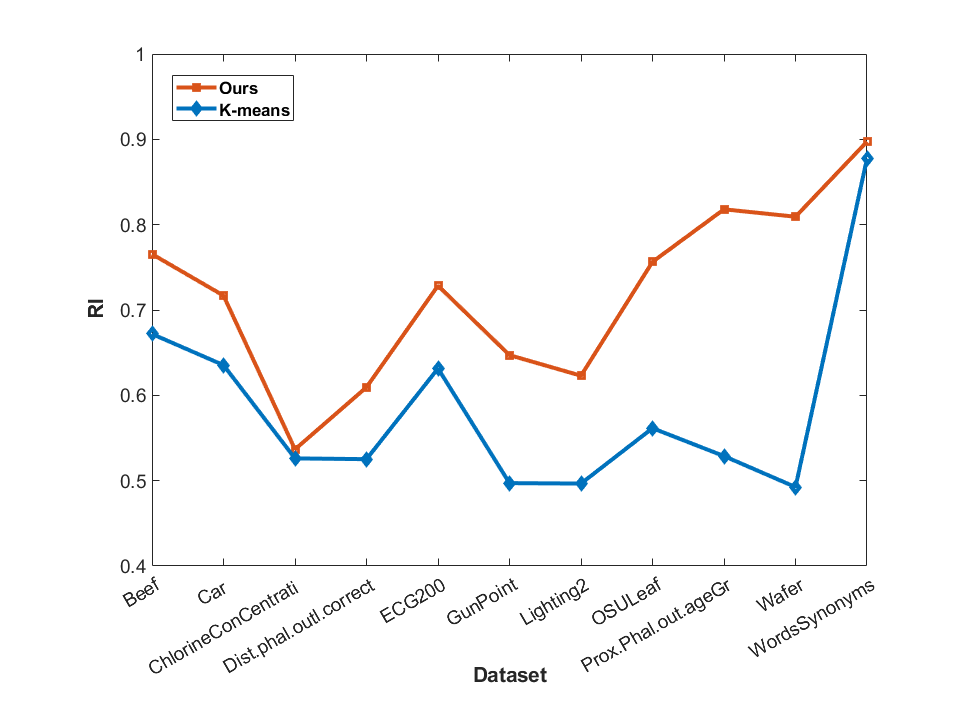}%fig2文件夹下的xbee.esp图片，
\caption{The RI values obtained by the K-means algorithm and ours.}
\end{figure}

On the contrary, our unsupervised clustering simply adopts an auto-encoding structure to update the parameters of our model and utilizes a K-means algorithm to classify the features obtained by RTFN. Although it is quite simple in structure, ours achieves decent performance on the 11 datasets, which depends on the strong feature extraction ability of RTFN.

In order to further testify that the proposed networks contribute to the accuracy improvement of the K-means algorithm, we compare the RTFN-based unsupervised clustering with a separate K-means algorithm on the 11 datasets above and show the RI results in Fig. 5. One can see that ours outperforms the separate K-means algorithm on each dataset. Sometimes, RI can be significantly improved, such as, on Wafer, Prox.Phal.out.ageGr and OSULeaf datasets. This is because our unsupervised clustering is provided with sufficient features obtained by the proposed RTFN, especially those hiding deeply in the input data which are beyond the exploration abilities of ordinary feature extraction networks. The results of the AVG ranks of all algorithms for comparison are shown in Fig. 6.

\begin{figure}[h]
  \centering
\includegraphics[height=3cm,width=14cm]{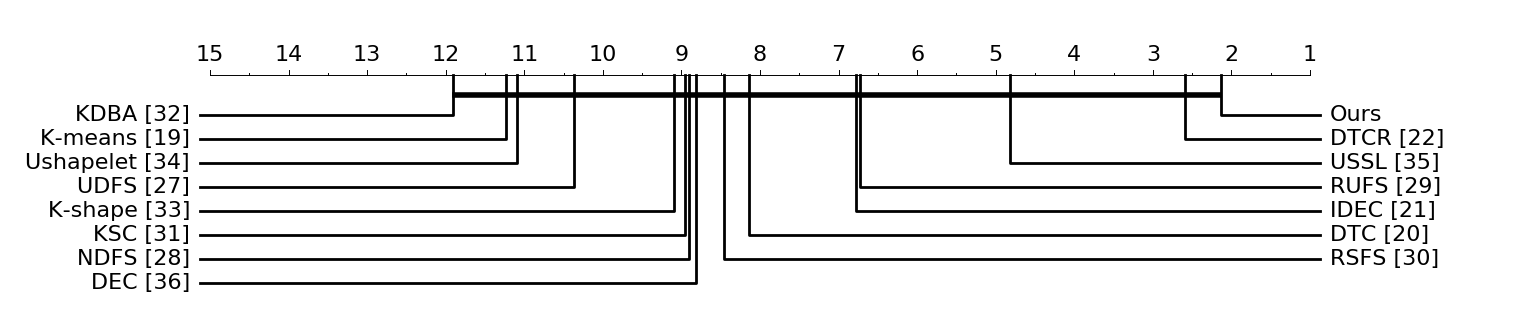}%fig2文件夹下的xbee.esp图片，
\caption{Results of the AVG ranks of the 15 algorithms.}
\end{figure}

\section{Conclusions}

In the proposed RTFN, the temporal feature networks are responsible for extracting basic features while the attentional LSTM networks target on those hidden shapelets and relationships which are extremely difficult to mine. Experimental results demonstrate that our RTFN achieves decent performance in both supervised classification and unsupervised clustering. Specifically, our RTFN-based supervised algorithm performs the best on 40 datasets, compared with the latest results from the supervised classification community. In particular, ours wins on 8 out of 12 long time-series datasets. Our RTFN-based unsupervised algorithm overweighs 14 state-of-the-art unsupervised algorithms on 11 datasets. Last but not least, the experimental results also indicate that RTFN has good potential to be embedded into other learning frameworks to handle time-series problems of various domains in the real world.

\section*{Broader Impact}
RTFN is a feature extraction network used to mine sufficient features from raw data. The features extracted by RTFN are beneficial to both supervised classification and unsupervised clustering, especially in time series classification. The high accuracy brought by RTFN facilitates development of useful applications in our daily life. For example, in the medical health domain, accurate ECG recognition improves clinical work efficiency of doctors and health professionals, promoting the development of AI medical care. In the earthquake warning domain, accurate classification is of great help for earthquake prediction, which saves human lives and reduces property losses. In the wildlife conservation domain, precise trajectory prediction also helps us to observe activity and behavior patterns of wild animals, which is in favor of their protection. On the other hand, due to its high complexity, RTFN cannot be directly deployed on mobile devices now, such as, mobile phones or intelligent watches. In addition, we do not think RTFN brings any negative impact in terms of its ethical aspects and future societal consequences.
\section*{References}
\small
[1] H. A. Dau, A. Bagnall, K. Kamgar, C.-C. M. Yeh, Y. Zhu, S. Gharghabi, C. A. Ratanamahatana and E. Keogh. The UCR time series archive. {\it arXiv preprint arXiv:1810.07758v2}, 2019.

[2] M. Längkvist, L. Karlsson and A. Loutfi. A review of unsupervised feature learning and deep learning for time-series modeling. {\it Pattern Recogn. Lett.}, vol. 42, no. 1, pp. 11-24, 2014.

[3] H. I. Fawaz, G. Forestier, J. Weber, L. Idoumghar and P. A. Muller. Deep learning for time series classification: a reviewer. {\it Data Min. Knowl. Disc.}, vol. 33, pp. 917-963, 2019.

[4] S. H. Huang, L. Xu and C. Jiang. Residual attention net for superior cross-domain time sequence modeling. {\it arXiv preprint arXiv: 2001.04077v1}, 2020.

[5] J. Han, H. Liu, M. Wang, Z. Li and Y. Zhang. ERA-LSTM: an efficient ReRAM-based architecture for long short-term memory. {\it IEEE Trans. Parallel Distrib. Syst.}, vol. 31, no. 6, pp. 1328-1342, 2020.

[6] F. Karim, S. Majumdar and H. Darabi. Insights into LSTM fully convolutional networks for time series classification. {\it IEEE Access}, vol. 7, pp. 67718-67725, 2019.

[7] M. Hajiaghayi and E. Vahedi. Code failure prediction and pattern extraction using LSTM networks. In {\it Proc. IEEE ICBDSC 2019}, Newark, USA, pp. 55-62, 2019.

[8] K. He, X. Zhang, S. Ren and J. Sun. Deep residual learning for image recognition. In {\it Proc. IEEE CVPR 2016}, Las Vegas, USA, pp. 770-778, 2016.

[9] C. Szegedy, W. Liu, Y. Jia, P. Sermanet, S. Reed, D. Anguelov, D. Erhan, V. Vanhoucke and A. Rabinovich. Going deeper with convolutions. In {\it Proc. IEEE CVPR 2015}, Boston, USA, pp. 7-12, 2015.

[10] A. Vaswani, N. Shazeer, N. Parmar, J. Uszkoreit, L. Jones, A. N. Gomez, Ł. Kaiser and I. Polosukhin. Attention is all you need. In {\it Proc. Adv. Neural Inf. Process. Syst. (ANIPS 2017)}, pp. 5998-6008, 2017.

[11] Y. Zhu, C. Zhao, H. Guo, J. Wang, X. Zhao and H. Lu. Attention couplenet: fully convolutional attention coupling network for object detection. {\it IEEE Trans. Image Process.}, vol. 28, no. 1, pp. 1170-1175, 2018.

[12] K. Xu, J. Ba, R. Kiros, K. Cho, A. Courville, R. Salakhutdinov, R. Zemel and Y. Bengio. Show, attend and tell: neural image caption generation with visual attention. In {\it Proc. ICML 2015}, Lille, France, pp. 2048-2057, 2015.

[13] H. Zhang, Z. Xiao, J. Wang, F. Li and E. Szczerbicki. A novel IoT-perceptive human activity recognition (HAR) approach using multi-head convolutional attention. {\it IEEE Internet Things J.}, vol. 2, no. 7, pp. 1072-1080, 2020.

[14] J. Chorowski, D. Bahdanau, D. Serdyuk, K. Cho and Y. Bengio. Attention-based models for speech recognition. In {\it Proc. Adv Neural Inf. Process. Syst. (ANIPS 2015)}, pp. 577-585, 2015.

[15] Z. Yang, X. He, J. Gao, L. Deng and A. Smola. Stacked attention networks for image question answering. In {\it Proc. IEEE CVPR 2016}, Las Vegas, USA, pp. 21-29, 2016.

[16] J. Chen, Z. Xiao, H. Xing, P. Dai, S. Luo and M. A. Lqbal. STDPG: a spatio-temporal deterministic policy gradient agent for dynamic routing in SDN. {\it arXiv preprint arXiv: 2004.098783}, 2020.

[17] S. Hochreiter and J. Schmihuber. Long short-term memory. {\it Neural Comput. Appl.}, vol. 9, no. 8, pp. 1735-1780, 1997.

[18] F. Karim, S. Majumdar, H. Darabi and S. Harford. Multivariate LSTM-FCNs for time series classification. {\it Neural Networks}, no. 116, pp. 237-245, 2019.

[19] J. A. Hartigan and M. A. Wong. Algorithm as 136: a K-means clustering algorithm. {\it J. Roy. Statist. Soc. Ser.} C 28, vol. 1, no. 28, pp. 100-108, 1979.

[20] N. S. Madiraju, S. M. Sadat, D. Fisher and H. Karimabadi. Deep temporal clustering: fully unsupervised learning of time-domain features. {\it arXiv preprint arXiv:1802.01059}, 2018.

[21] X. Guo, L. Gao, X. Liu and J. Yin. Improved deep embedded clustering with local structure preservation. In {\it Proc. IJCAI 2017}, Melbourne, Australia, pp. 1753-1759, 2017.

[22] Q. Ma, J. Zheng, S. Li and G. W. Cottrell. Learning representations for time series clustering. In {\it Proc. NeurIPS 2019}, Vancouver, Canada, 2019.

[23] J. Large, A. Bagnall, S. Malinowski and R. Tavenard. From BOP to BOSS and beyond: time series classification with dictionary based classifier. {\it arXiv preprint arXiv:1809.06751}, 2018.

[24] A. Bagnall, J. Lines, A. Bostrom, J. Large and E. Keogh. The great time series classification bake of: a review and experimental evaluation for recent algorithmic advances. {\it Data Min. Knowl. Disc.}, vol. 31, pp. 606-660, 2017.

[25] J. Lines and A. Bagnall. Time series classification with ensembles of elastic distance measures. {\it  Data Min. Knowl. Disc.}, vol. 29, pp. 565-592, 2015.

[26] W. M. Rand. Objective criteria for the evaluation of clustering methods. {\it J. Am. Stat. Assoc.}, vol. 166, no. 336, pp. 845-850, 1971.

[27] Y. Yang, H. Shen, Z. Ma, Z. Huang and X. Zhou. L2, 1-norm regularized discriminative feature selection for unsupervised. In {\it Proc. IJCAI 2011}, Barcelona, Spain, pp. 1589-1594, 2011.

[28] Z. Li, Y. Yang, J. Liu, X. Zhou and H. Lu. Unsupervised feature selection using nonnegative spectral analysis. In {\it Proc. AAAI 2012}, Toronto, Canada, pp. 1026-1032, 2012.

[29] M. Qian and C. Zhai. Robust unsupervised feature selection. In {\it Proc. IJCAI 2013}, Beijing, China, pp. 1621-1627, 2013.

[30] L. Shi, L. Du and Y. Shen. Robust spectral learning for unsupervised feature selection. In {\it Proc. IEEE ICDM 2014}, Shenzhen, China, pp. 977-982, 2014.

[31] J. Yang and J. Leskovec. Patterns of temporal variation in online media. In {\it Proc. ACM WSDM 2011}, Hong Kong, pp. 177-186, 2011.

[32] F. Petitjean, A. Ketterlin and P. Gancarski. A global averaging method for dynamic time warping, with the applications to clustering. {\it Pattern Recogn.}, vol. 44, no. 3, pp. 678-693, 2011.

[33] J. Paparrizos and L. Gravano. K-shape: efficient and accurate clustering of time series. In {\it Proc. ACM SIGMOD 2015}, Melbourne, Australia, pp. 1855-1870, 2015.

[34] J. Zakaris, A. Mueen and E. Keogh. Clustering time series using unsupervised shapelets. In {\it Proc. IEEE ICDM 2012 }, Brussels, Belgium, pp. 785-794, 2012.

[35] Q. Zhang, J. Wu, P. Zhang, G. Long and C. Zhang. Salient subsequence learning for time series clustering. {\it IEEE Trans. Pattern Anal.}, vol 41, no. 9, pp. 2193-2207, 2018.

[36] J. Xie, R. Girshick and A. Farhadi. Unsupervised deep embedding for clustering analysis. In {\it Proc. ICML 2016}, New York, USA, pp. 478-487, 2016.

\clearpage
\section*{Appendix}

\begin{table}[!htb]
  \caption{The original results of the top-1 accuracy obtained from 85 datasets.}
  \label{sample-table}
  \centering
  \resizebox{140mm}{95mm}{
  \begin{tabular}{cccccccc }
    \toprule

     Dataset  & Existing SOTA {[}4{]} & Best:lstm-fcn & Vanilla:ResNet-Transformer & ResNet-Transformer 1 {[}4{]} & ResNet-Transformer 2 {[}4{]} & ResNet-Transformer 3 {[}4{]} & Ours \\
       \midrule

Adiac	&0.857	&\textbf{0.869565}	&0.84399&	0.849105&	0.849105	&0.849105	&0.792839\\
ArrowHead	&0.88	&\textbf{0.925714}	&0.891429	&0.891429	&0.891429	&0.891429	&0.851429\\
Beef&	\textbf{0.9}	&\textbf{0.9}&	0.866667&	0.866667&	0.8666667&	0.8666667&	\textbf{0.9}\\
BeetleFly	&0.95	&\textbf{1}	&\textbf{1}	&0.95	&0.95	&\textbf{1}&	\textbf{1}\\
BirdChicken&	0.95&	0.95&	\textbf{1}&	0.9&	\textbf{1}&	0.7&\textbf{	1}\\
Car&	0.933&	\textbf{0.966667}&	0.95	&0.883333	&0.866667	&0.3&	0.883333\\
CBF&	\textbf{1}&	0.996667&	\textbf{1}&	0.997778&	\textbf{1}&	\textbf{1}&	\textbf{1}\\
ChlorineConcentration&	0.872	&0.816146&	0.849479&	0.863281&	0.409375&	0.861719&	\textbf{0.894271}\\
CinCECGTorso&	\textbf{0.9949}&	0.904348&	0.871739&	0.656522&	0.89058&	0.31087&	0.810145\\
Coffee&\textbf{1}	&\textbf{1}	&\textbf{1}	&\textbf{1}	&\textbf{1}	&\textbf{1}	&\textbf{1}\\
CricketX	&0.821&	0.792308	&\textbf{0.838462}	&0.8&	0.810256&	0.8&	0.771795\\
CricketY	&0.8256&	0.802564&	\textbf{0.838462}&	0.805128&	0.825641&	0.808766&	0.789744\\
CricketZ	&0.8154&	0.807692&	\textbf{0.820513}&	0.805128&	0.128205&	0.1&	0.787179\\
DiatomSizeReduction&	0.967&	0.970588&	0.993464&	\textbf{0.996732}&	0.379085&	\textbf{0.996732}&	0.980392\\
DistalPhalanxOutlineAgeGroup&	\textbf{0.835}&	0.791367&	0.81295&	0.776978&	0.467626&	0.776978&	0.719425\\
DistalPhalanxOutlineCorrect&	0.82&	0.791367&	\textbf{0.822464}	&\textbf{0.822464}	&\textbf{0.822464}	&0.793478&	0.771739\\
Earthquakes	&0.801	&\textbf{0.81295}&	0.755396&	0.755396&	0.76259	&0.755396	&0.776978\\
ECG200	&0.92	&0.91&	0.94&	\textbf{0.95}&	0.94&	0.93&	0.92\\
ECG5000	&0.9482	&\textbf{0.948222}&	0.941556&	0.943556&	0.944222&	0.940444&	0.944444\\
ECGFiveDays	&\textbf{1}&	0.987224&	\textbf{1}&	\textbf{1}&	\textbf{1}&\textbf{	1}&	0.954704\\
FaceFour&	\textbf{1}&	0.943182&	0.954545&	0.965909&	0.977273&	0.215909&	0.924045\\
FacesUCR	&\textbf{0.958}&	0.941463&	0.957561&	0.947805&	0.926829&	0.95122&	0.870244\\
FordA	&0.9727	&\textbf{0.976515}&	0.948485&	0.946212&	0.517424&	0.940909&	0.939394\\
FordB	&\textbf{0.9173}	&0.792593	&0.838272	&0.830864&	0.838272&	0.823457&	0.823547\\
GunPoint	&\textbf{1}	&\textbf{1}&	\textbf{1}&	\textbf{1}&	\textbf{1}&	\textbf{1}&	\textbf{1}\\
Ham	&0.781&	\textbf{0.809524}&	0.761905&	0.780952&	0.619048&	0.514286&	\textbf{0.809524}\\
HandOutlines&	0.9487&	\textbf{0.954054}&	0.937838&	0.948649&	0.835135&	0.945946&	0.894595\\
Haptics&	0.551&	0.558442	&0.564935	&0.545455	&\textbf{0.600649}&	0.194805	&\textbf{0.600649}\\
Herring&0.703	&\textbf{0.75}	&0.703125	&0.734375	&0.65625	&0.703125	&\textbf{0.75}\\
InsectWingbeatSound&	0.6525	&\textbf{0.668687}	&0.522222	&0.642424&0.53859&0.536364	&0.651515\\
ItalyPowerDemand&0.97&0.963071&	0.965015&	0.969874&	0.962099&	\textbf{0.971817}&	0.964043\\
Lightning2&\textbf{0.8853}&	0.819672&	0.852459&	0.852459&	0.754098&	0.868852&	0.836066\\
Lightning7	&0.863&	0.863014&	0.821918&	0.849315&	0.383562&	0.835616&	\textbf{0.90411}\\
Mallat&	0.98&	\textbf{0.98081}&	0.977399&	0.975267&	0.934328&	0.979104&	0.938593\\
Meat	&\textbf{1}	&0.883333&	\textbf{1}&	\textbf{1}&	\textbf{1}&	\textbf{1}&\textbf{	1}\\
MedicalImages	&0.792&	\textbf{0.798684}&	0.780263&	0.765789&	0.759211&	0.789474&	0.793421\\
MiddlePhalanxOutlineAgeGroup&	\textbf{0.8144}&	0.668831&	0.655844&	0.662338&	0.623377&	0.662338&	0.662388\\
MiddlePhalanxOutlineCorrect&0.8076&	0.841924&	\textbf{0.848797}&	\textbf{0.848797}&	\textbf{0.848797}&	0.835052&	0.744755\\
MiddlePhalanxTW&	0.612&	0.603896&	0.564935&	0.577922&	0.551948&	\textbf{0.623377}&	\textbf{0.623377}\\
MoteStrain&\textbf{0.95}&0.938498&	0.940895&	0.916933&	0.9377&	0.679712&	0.875399\\
OliveOil&	0.9333&	0.766667&	\textbf{0.966667}&	0.9&	0.933333&	0.9&\textbf{	0.966667}\\
Plane	&\textbf{1}	&\textbf{1}	&\textbf{1}	&\textbf{1}	&\textbf{1}	&\textbf{1}	&\textbf{1}\\
ProximalPhalanxOutlineAgeGroup	&0.8832&	0.887805&	0.887805&	\textbf{0.892683}&	0.882927&	\textbf{0.892683}&	0.878049\\
ProximalPhalanxOutlineCorrect&	0.918&	\textbf{0.931271}&\textbf{	0.931271}&	\textbf{0.931271}&	0.683849&	0.924399&	0.910653\\
ProximalPhalanxTW&	0.815&\textbf{	0.843902}	&0.819512&	0.814634&	0.819512&	0.819512&	0.834146\\
ShapeletSim&	\textbf{1}&	\textbf{1}&\textbf{	1}&	0.91111&	0.888889&	0.9777778&\textbf{	1}\\
ShapesAll	&0.9183&	0.905&	0.923333&	0.876667&	0.921667&\textbf{	0.933333}&	0.876667\\
SonyAIBORobotSurface1&	0.985&	0.980525&	\textbf{0.988353}&	0.978369&	0.708819&	0.985025&	0.881864\\
SonyAIBORobotSurface2&	0.962&	0.972718	&0.976915&	0.974816	&\textbf{0.98426}&	0.976915&	0.854145\\
Strawberry	&0.976&	\textbf{0.986486}&\textbf{	0.986486}&	\textbf{0.986486}&	\textbf{0.986486}&	\textbf{0.986486}	&\textbf{0.986486}\\
SwedishLeaf	&0.9664&	\textbf{0.9792}&	\textbf{0.9792}&	0.9728&	0.9696&	0.9664&	0.9376\\
Symbols&	0.9668&\textbf{	0.98794}&	0.9799&	0.970854&	0.976884&	0.252261&	0.892462\\
SyntheticControl	&\textbf{1}&	0.993333&	\textbf{1}&	0.996667&	\textbf{1}&\textbf{	1}&	\textbf{1}\\
ToeSegmentation1&	0.9737&	\textbf{0.991228}&	0.969298&	0.969298&	0.97807&	\textbf{0.991228}&	0.982456\\
ToeSegmentation2&	0.9615&	0.930769&	\textbf{0.976923}&	0.953846&	0.953846&	\textbf{0.976923}&	0.938462\\
Trace	&\textbf{1}&	\textbf{1}&\textbf{	1}&\textbf{	1}&\textbf{	1}&\textbf{	1}&\textbf{	1}\\
TwoLeadECG	&\textbf{1}&	\textbf{1}&\textbf{	1}&\textbf{	1}&\textbf{	1}&\textbf{	1}&\textbf{	1}\\
TwoPatterns	&\textbf{1}	&0.99675	&\textbf{1}	&\textbf{1}	&\textbf{1}	&\textbf{1}	&\textbf{1}\\
UWaveGestureLibraryAll	&\textbf{0.9685}&	0.961195&	0.856784&	0.933277&	0.939978&	0.879118&	\textbf{0.9685}\\
UWaveGestureLibraryX	&0.8308	&\textbf{0.843663}	&0.780849&	0.814629&	0.810999&	0.808766&	0.815187\\
UWaveGestureLibraryY	&0.7585&	\textbf{0.765215}&	0.664992&	0.71636&	0.671413&	0.67895&	0.752094\\
UWaveGestureLibraryZ	&0.7725&	\textbf{0.795924}&	0.756002&	0.761027&	0.760469&	0.762144&	0.757677\\
Wafer	&\textbf{1}&	0.998378&	0.99854&	0.998215&	0.99854&	0.999027&	\textbf{1}\\
Wine	&0.889&	0.833333&	0.851852&	0.87037&	0.87037&	\textbf{0.907407}&	\textbf{0.907407}\\
WordSynonyms	&0.779&	\textbf{0.680251}&	0.661442&	0.65047&	0.636364&	0.678683&	0.659875\\
ACSF1	&---	&0.9	&\textbf{0.96}&	0.91&	0.93&	0.17	&0.83\\
BME	&---&	0.993333&	\textbf{1}&\textbf{	1}&\textbf{	1}&\textbf{	1}&	0.986667\\
Chinatown&	---&	0.982609&	\textbf{0.985507}&\textbf{	0.985507}&\textbf{	0.985507}&	\textbf{0.985507}	&\textbf{0.985507}\\
Crop	&---&	0.74494&	0.743869&	0.742738&	0.746012&	0.740476&	\textbf{0.756071}\\
DodgerLoopDay&	---&	0.6375&	0.5357&	0.55&	0.4625&	0.5&\textbf{	0.675}\\
DodgerLoopGame	&---&	0.898551&	0.876812&	0.891304&	0.550725&	\textbf{0.905797}&	\textbf{0.905797}\\
DodgerLoopWeekend&	---	&0.978261&	0.963768	&0.978261&	0.949275&	0.963768&	\textbf{0.985507}\\
GunPointAgeSpan	&---&	0.996835&	0.996835&	0.996835&	\textbf{1}&	0.848101&	0.990506\\
GunPointMaleVersusFemale&	---&	\textbf{1}&\textbf{	1}&\textbf{	1}&	0.996835&	0.996835&	\textbf{1}\\
GunPointOldVersusYoung	&---&	0.993651&	\textbf{1}&\textbf{	1}&\textbf{	1}&\textbf{	1}&\textbf{	1}\\
InsectEPGRegularTrain&	---&	0.995984&	\textbf{1}&	\textbf{1}&\textbf{	1}&\textbf{	1}&	\textbf{1}\\
InsectEPGSmallTrain&	---&	0.935743&	0.955823&	0.927711&	0.971888&	0.477912&	\textbf{1}\\
MelbournePedestrian&	---&	0.913061&	0.912245&	0.911837&	0.904898&	0.901633&	\textbf{0.95736}\\
PowerCons	&---&	0.994444&	0.933333&	0.9444444&	0.927778&	0.927778&	\textbf{1}\\
Rock	&---&	\textbf{0.92}&	0.78&	\textbf{0.92}&	0.82&	0.76&	0.88\\
SemgHandGenderCh2&	---&	0.91	&0.866667&	0.916667&	0.848333	&0.651667&	\textbf{0.923333}\\
SemgHandMovementCh2&	---&	0.56	&0.513333&	0.504444	&0.391111	&0.468889&	\textbf{0.757778}\\
SemgHandSubjectCh2	&---&	0.873333&	0.746667&	0.74&	0.666667&	0.788889&	\textbf{0.897778}\\
SmoothSubspace	&---&	0.98	&\textbf{1}	&\textbf{1	}&0.99333&	\textbf{1}	&\textbf{1}\\
UMD	&---&	0.986111&	\textbf{1}&\textbf{	1}&	\textbf{1}&\textbf{	1}&	\textbf{1}\\

    \bottomrule
  \end{tabular}
  }
\end{table}

\end{document}